\documentclass{article}

\usepackage[numbers]{natbib}

     \usepackage[preprint]{neurips_2025}

\usepackage[utf8]{inputenc} % allow utf-8 input
\usepackage[T1]{fontenc}    % use 8-bit T1 fonts
\usepackage{hyperref}       % hyperlinks
\usepackage{url}            % simple URL typesetting
\usepackage{booktabs}       % professional-quality tables
\usepackage{amsfonts}       % blackboard math symbols
\usepackage{nicefrac}       % compact symbols for 1/2, etc.
\usepackage{microtype}      % microtypography
\usepackage{xcolor}         % colors
\usepackage{graphicx}
\usepackage{amsmath}
\usepackage[normalem]{ulem}
\usepackage{enumitem}

\usepackage{multirow}
\usepackage{wrapfig}
\usepackage{subfig}
\usepackage{amsthm}
\usepackage{xcolor}
\usepackage{tcolorbox}

\title{Beyond path selection: Better LLMs  for Scientific Information Extraction with MimicSFT and Relevance and Rule-induced(R$^2$)GRPO}

\author{%
    Ran Li \\
    HKUST\\
    Hong Kong SAR, China \\
    \texttt{rlibb@connect.ust.hk} \\
    \And
    Shimin Di \\
    SEU\\
    Jiangsu, China \\
    \texttt{shimin.di@seu.edu.cn} \\
    \And
    Yuchen Liu \\
    HKUST(GZ)\\
    Guangzhou, China \\
    \texttt{yliu356@connect.hkust-gz.edu.cn} \\
    \And
    Chen Jing \\
    Zhipu AI\\
    Beijing, China \\
    \texttt{chen.jing@aminer.cn} \\
    \And
    Yu Qiu \\
    Zhipu AI\\
    Beijing, China \\
    \texttt{yu.qiu@aminer.cn} \\
    \And
    Lei Chen \\
    HKUST(GZ), HKUST\\
    Guangzhou, China \\
    \texttt{leichen@hkust-gz.edu.cn} \\
}
\begin{document}

	\maketitle

	\begin{abstract}
		Previous study suggest that powerful Large Language Models (LLMs) trained with Reinforcement Learning with Verifiable Rewards (RLVR) only refines reasoning path without improving the reasoning capacity in math tasks while supervised-finetuning(SFT) with distillation can. We study this from the view of Scientific information extraction (SciIE) where LLMs and reasoning LLMs underperforms small Bert-based models. SciIE require both the reasoning and memorization. We argue that both SFT and RLVR can refine the reasoning path and improve reasoning capacity in a simple way based on SciIE. We propose two-stage training with 1. MimicSFT, using structured reasoning templates without needing high-quality chain-of-thought data, 2. R$^2$GRPO with relevance and rule-induced rewards. Experiments on scientific IE benchmarks show that both methods can improve the reasoning capacity. R$^2$GRPO with mimicSFT surpasses baseline LLMs and specialized supervised models in relation extraction.
		Our code is available at \url{https://github.com/ranlislz/R2GRPO} .
	\end{abstract}
	\section{Introduction}
	\label{sec:introduction}

	Reasoning Large Language Models (LLMs) \cite{guo2025deepseek,jaech2024openai, el2025competitive}, trained with Reinforcement Learning from Verifiable Rewards (RLVR), have shown complex tasks like mathematical reasoning and code generation. By integrating chain-of-thought (CoT) prompting with RL-driven path optimization, models such as DeepSeek-R1 \cite{guo2025deepseek}, OpenAI O1,O3 \cite{el2025competitive,jaech2024openai} and Google gemini-2.0 thinking \cite{gemini2.0-think} iteratively refine their reasoning trajectories, achieving human-level performance in select domains. 

\begin{figure}[!t]
	\vspace{-5pt}
	\centering
	\includegraphics[width=0.98\textwidth]{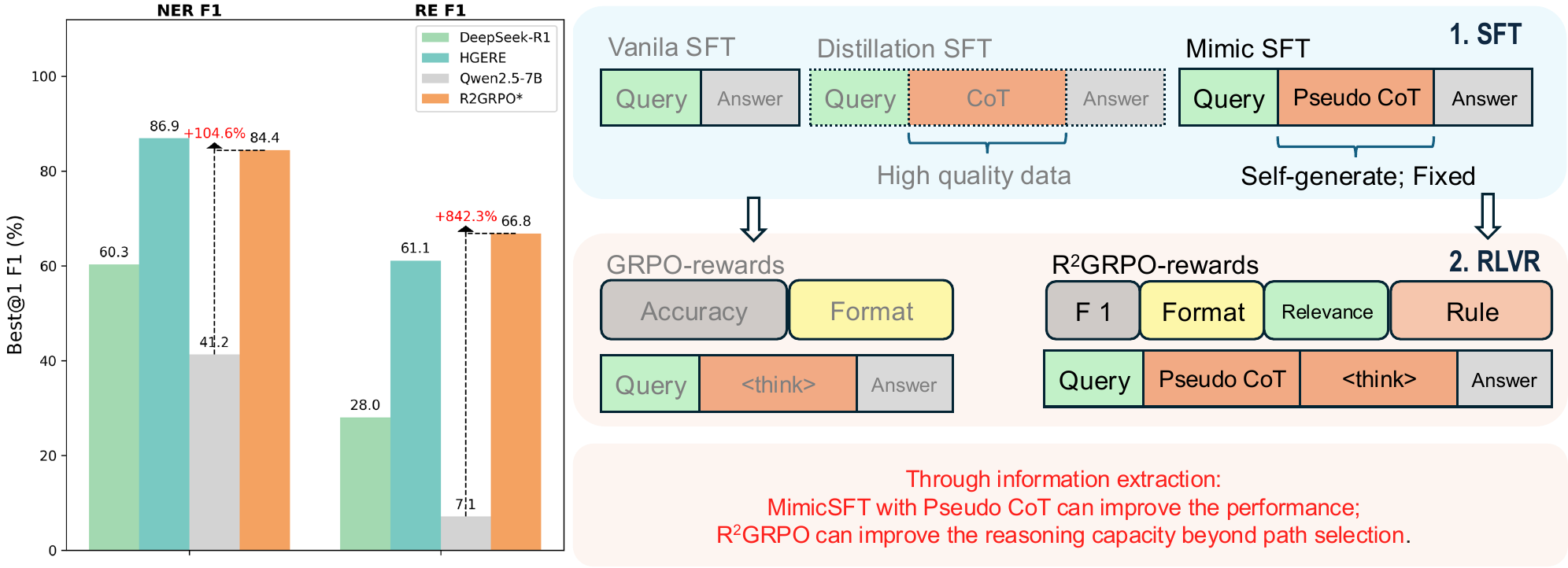}
	\vspace{-5pt}
	\caption{
		Our two-stage training for scientific IE (right) and the performance gain (left) }
	\label{fig:overview}
	\vspace{-1pt}
\end{figure}
	On the one hand, recent studies suggest RLVR optimizes output distributions rather than imparting new abilities. For instance, RL-tuned LLMs exhibit improved pass@1 rates on math benchmarks but reduced pass@k diversity, implying narrower path selection instead of enriched reasoning capacity \cite{yue2025does}. Conversely, supervised fine-tuning (SFT) with knowledge distillation \cite{distillinghinton2015}from larger models has been shown to enhance reasoning breadth \cite{yue2025does,guo2025deepseek}.
	On the other hand, the less explored Information extraction (IE) for reasoning LLMs presents an ideal task for investigating this question. As shown in Figure~\ref{fig:overview}, even state-of-the-art LLMs underperform supervised BERT-based models \cite{devlin2018bert, beltagy2019scibert, yan2023joint, ye2021packed,zhong2020frustratingly} on Scientific IE(SciIE) benchmarks like SciER \cite{zhang2024scier}. 
	The task demands precise recall of domain-specific entities (memorization) and systematic reasoning to infer implicit relations. LLMs often perform bad on entity span detection and relational inference, highlighting a misalignment between their training objectives and IE's dual requirements. 
	
	We argue that IE's hybrid nature with both \textit{knowledge memorization} and \textit{contextual rule reasoning} makes it an ideal lens to study what RLVR truly learns. \cite{chu2025sft} States SFT is better at memorization and RL is good at generalization. However, \cite{shao2024deepseekmath} provide the unified representation of SFT and RLVR for LLM post-training. From this view, both RLVR and SFT should be able to conduct memorization and optimize reasoning in a simple way.
	Based on this, it is also interesting if we can improve the LLMs performance to be comparable with the supervised models.
\begin{figure}[!t]
	\centering
	\includegraphics[width=0.95\textwidth]{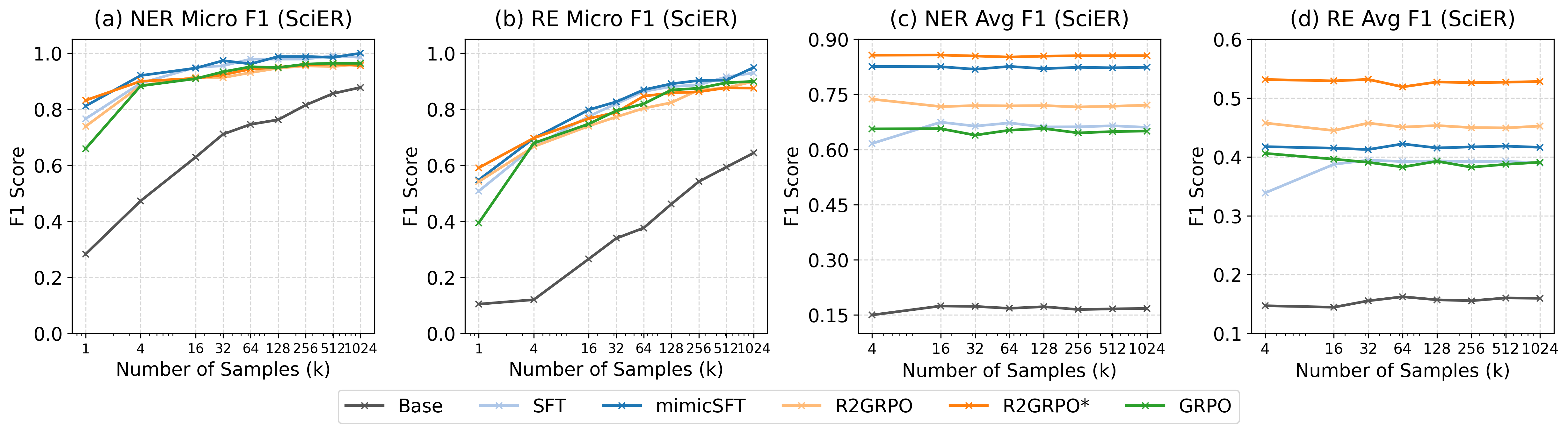}
	\vspace{-5pt}
	\caption{Best F1@K scores representing the reasoning capacity and Avg@K scores representing the reasoning ability for NER and RE on SciER (small).}
	\label{fig:bestk_max_scier}
	\vspace{-10pt}
\end{figure}

\begin{figure}[!t]
	\centering
	\includegraphics[width=0.95\textwidth]{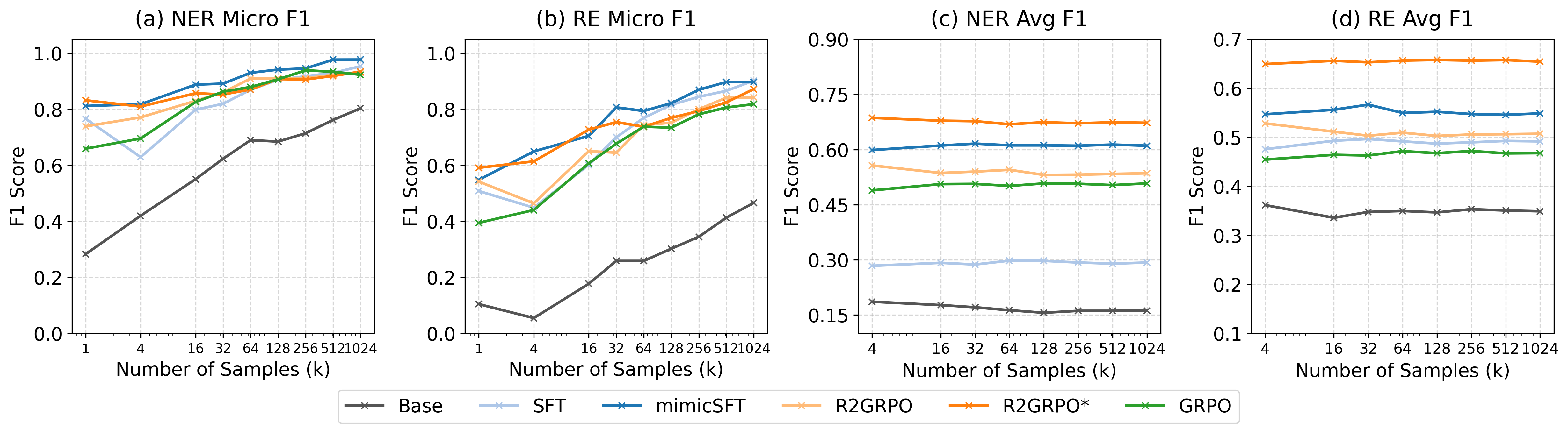}
	\vspace{-5pt}
	\caption{Best F1@K scores representing the reasoning capacity and Avg@K scores representing the reasoning ability for NER and RE on OOD (small).}
	\vspace{-10pt}
	\label{fig:bestk_ood}
\end{figure}
	
	Our systematic study of RLVR's impact on IE reveals that it can enhance both knowledge retention and reasoning when applied under task-aware conditions. Contrary to prior assumptions, RLVR and SFT exhibit complementary effects: RLVR improves both Best@1 and Best@k performance, suggesting it enhances both reasoning ability and capacity\cite{yue2025does}. 
	Furthermore, we discover that SFT with structured reasoning templates (which we call MimicSFT) can significantly boost performance without requiring high-quality CoT data, challenging previous view on math datasets \cite{yue2025does}. We noticed that for the reasoning, the structure and relevance are more important than the length for a constrained generation problem.
	Based on these insights, we propose \textbf{R\textsuperscript{2}GRPO}, a two-phase training method that combines MimicSFT with a novel reward function incorporating relevance and rule-mining to jointly optimize knowledge grounding and reasoning paths. Our hierarchical reasoning approach decomposes complex SciIE tasks into more tractable sub-problems, guiding the model through structured reasoning steps that satisfy both schema and factual constraints. Our key contributions are:
	\begin{itemize}[leftmargin=*]
	\item We show RLVR can enhance both knowledge memorization and systematic reasoning in LLMs for SciIE, challenging the view that RLVR only refines reasoning path.
	\item Further we propose MimicSFT, with pseudo reasoning templates, a simple adaptation without requiring high-quality CoT data can improve the reasoning ability of models on SciIE.
	\item The development of R\textsuperscript{2}GRPO, which integrates MimicSFT with a composite reward function encouraging relevance and rule induction during reasoning, achieving state-of-the-art IE performance among reasoning models and comparable with supervised models.
\end{itemize}

	\section{Related Work}
	\label{sec:related_work}
	
	\textbf{Post-training LLMs for Enhanced Capabilities.} Large Language Models \cite{gpt4achiam2023} pre-trained on vast data, acquiring abundant knowledge and diverse abilities \cite{scalingkaplan2020}. However, aligning these models with specific downstream tasks or desired behaviors often requires post-training \cite{optzhang2022,hoffmann2022training,geminiteam2024, touvron2023llama}, primarily Supervised Fine-Tuning (SFT) \cite{finetuneradford2018improving,fewshotbrown2020language, wei2021finetuned, chung2024scaling, zhou2023lima}and Reinforcement Learning (RL) \cite{ziegler2019fine,ouyang2022training, guo2025deepseek, shao2024deepseekmath}. Post-training adapts models to specific input-output formats using labeled examples, effectively injecting task-specific knowledge or styles \cite{ouyang2022training,bai2022training}.Chain-of-Thought prompting \cite{wei2022chain} and RLVR methods like GRPO \cite{shao2024deepseekmath} encouraging models to self-generate the reasoning chain have significantly boosted LLMs' performance on reasoning tasks like mathematics and code generation.  \cite{chu2025sft} states that SFT is mainly for memorization and  RL can improve generalization.  However, \cite{yue2025does} argue that RLVRs mainly optimize the reasoning paths that the base model already has with its pre-trained knowledge. \cite{li2502llms} mentions the structured reasoning pattern is more important than the content. \cite{sprague2024cot} shows that the CoT is good  for math and symbolic reasoning but not necessary for other types tasks. 
	On the other hand, some argue that SFT can also help to gain reasoning capacity \cite{li2024getting}.  
	So is there a clear gap between SFT and RLVR and what do RLVR really learns? Our work investigates this question within the less explored domain of Information Extraction, where both knowledge and reasoning are crucial.
	
	\textbf{Scientific Information Extraction (SciIE).} IE aims to automatically extract structured information, such as entities, relations, and events, from unstructured text. Traditional approaches often rely on supervised learning with sequence labeling models like BiLSTMs or Transformer-based architectures (e.g., BERT) trained on domain-specific annotated datasets ~\cite{devlin2018bert}. These methods achieve high performance on current benchmarks, particularly for Named Entity Recognition (NER) and Relation Extraction (RE), but require significant domain annotation data. Recently, LLMs have been explored for IE tasks, leveraging their zero-shot \cite{zerowei2023,revisitingli2023,lu2023pivoine, xie2023empirical,yuan2023zero} or in-context learning capabilities \cite{bi2024codekgc,zhang2024extract,zhu2024llms} or undergoing supervised fine-tuning \cite{wang2023gpt, dagdelen2024structured, ning2024urbankgent}. While LLMs offer flexibility, they often under-perform specialized supervised models \cite{zhong2020frustratingly,yan2023joint,ye2021packed} trained with domain knowledge, especially in scientific domains(SciIE) \cite{beltagy2019scibert,zhang2024scier, dagdelen2024structured} requiring specialized knowledge or complex relational inference. 
 Few studies have study how to adapt the reasoning ability of LLMs through post-training to improve the perfromance on SciIE, a gap our work aims to fill by examining how SFT and RL differentially contribute to performance on Scientific NER and RE.
	
	\section{Methodology}
	\label{sec:methodology}
	
	\subsection{Problem Formulation}
	\label{subsec:problem_formulation}
	We focus on two fundamental information extraction tasks: Named Entity Recognition (NER) and Relation Extraction (RE). 
	
	\textbf{Named Entity Recognition (NER):} Given an input text $\mathbf{x} = \{x_1, x_2, \ldots, x_n\}$, NER identifies entity spans $e_i = \{x_j, \ldots, x_k\}$ and assigns each a type $t_i \in \mathcal{T}$, where $\mathcal{T}$ is a predefined set of entity types (e.g., Task, Method, Dataset in scientific literature).
	
	\textbf{Relation Extraction (RE):} For a pair of entities $(e_i, e_j)$ identified in text, RE determines whether a relation exists and, if so, classifies it into a relation type $r_{ij} \in \mathcal{R}$, where $\mathcal{R}$ is the set of possible relation types (e.g., Used-For, Compare-With).
	
	\textbf{End-to-End IE:} This combines both tasks, requiring models to first identify entities and then determine relations between them, making it particularly challenging as errors in entity recognition propagate to relation extraction.
	
	\textbf{Constrained Generation View:} We can view IE as a constrained generation problem where the model must generate outputs $y$ that satisfy both:
	\begin{itemize}[leftmargin=*]
	    \item \textit{Schema constraints}: Answers must conform to predefined entity and relation types and follow the required structure (e.g. valid json format).
	    \item \textit{Factual constraints}: Answers must come from the original content. 
	\end{itemize}
	
	Formally, we can define the constrained generation problem as finding:
	\begin{equation}
	    y^* = \arg\max_{y \in \mathcal{Y}} P(y|x; \theta) \quad \text{s.t.,} \quad C_{\text{schema}}(y) = 1 \land C_{\text{factual}}(y, x) = 1,
	\end{equation}
	where $C_{\text{schema}}$ and $C_{\text{factual}}$ are binary constraint functions. This formulation is challenging for standard LLMs as they must simultaneously satisfy structural constraints while maintaining factual accuracy.
	All fine-tuning is performed using Low-Rank Adaptation (LoRA) \citep{hu2022lora} for computational efficiency.

	\subsection{Supervised Fine-Tuning and MimicSFT}
	\label{subsec:sft_exploration}
	
	Standard SFT adapts a pre-trained LLM by maximizing the conditional probability of target outputs given inputs:
	\begin{equation}
	    \label{eq:sft_loss}
	    \mathcal{L}_{\text{SFT}}(\theta) = - \sum_{(x, y) \in \mathcal{D}_{\text{SFT}}} \log P(y | x; \theta),
	\end{equation}
	where $\mathcal{D}_{\text{SFT}}$ is the supervised fine-tuning dataset. In terms of Equation \ref{eq:unified_gradient}, $o=y$, $\mathcal{D} = \mathcal{D}_{\text{SFT}}$, and $GC(x, y, t, \pi_{\text{ref}}) = 1$ for all tokens.
	
	To improve generalization, we decompose IE into distinct sub-tasks (NER only, RE with Gold Entities, RE only, End-to-End IE) and employ a multi-task learning approach:
	\begin{equation}
	    \label{eq:sft_mt_loss}
	    \mathcal{L}_{\text{MT-SFT}}(\theta) = - \sum_{k=1}^K \sum_{(x, y) \in \mathcal{D}_{\text{SFT}, T_k}} \log P(y | x, T_k; \theta)
	\end{equation}
	where $T_k$ indicates the task type in the prompt.
	
	\textbf{MimicSFT: Structured Reasoning Without CoT Data.} We introduce MimicSFT to encourage structured reasoning without requiring high-quality CoT annotations. The model is trained to produce a templated reasoning block $z$ (enclosed in \texttt{<reasoning>...</reasoning>} tags) before generating the final output $y$:
	\begin{equation}
	    \label{eq:mimic_sft_loss}
	    \mathcal{L}_{\text{MimicSFT}}(\theta) = - \sum_{(x, y') \in \mathcal{D}_{\text{MimicSFT}}} \log P(y' | x; \theta),
	\end{equation}
	where $y' = (z, y)$ is the concatenation of reasoning steps and final output. The reasoning template follows a general IE process (e.g., 1. Identify entities, 2. Consider relations, 3. Formulate extraction).

	\subsection{R$^2$GRPO: Reinforcement Learning with Relevance and Rule-Induction}
\label{subsec:r2_rl}

\subsubsection{GRPO Framework}
R$^2$GRPO builds on Group Relative Policy Optimization (GRPO) \citep{shao2024deepseekmath}, a PPO variant that normalizes rewards based on group performance. The GRPO objective is:
\begin{align}
	\label{eq:grpo_objective}
	J_{\text{GRPO}}(\theta) &= \mathbb{E}_{q \sim \mathcal{D}, \{o_i\}_{i=1}^G \sim \pi_{\theta_{\text{old}}}(O|q)} \Big[ \frac{1}{G} \sum_{i=1}^{G} \frac{1}{|o_i|} \sum_{t=1}^{|o_i|} \mathcal{A}_{\text{clip}}(o_{i,t}, q, \hat{A}_{i,t}) \notag \\
	&\quad - \beta D_{\text{KL}}(\pi_{\theta}(\cdot|q, o_{i,<t}) || \pi_{\text{ref}}(\cdot|q, o_{i,<t})) \Big],
\end{align}

where:
	, $q$ is an input prompt from the IE dataset
	, $\{o_i\}_{i=1}^G$ is a group of $G$ outputs sampled from policy $\pi_{\theta_{\text{old}}}$
	, $\mathcal{A}_{\text{clip}}(o_{i,t}, q, \hat{A}_{i,t}) = \min(r_t(\theta) \hat{A}_{i,t}, \text{clip}(r_t(\theta), 1-\epsilon, 1+\epsilon) \hat{A}_{i,t})$
	, $r_t(\theta) = \frac{\pi_{\theta}(o_{i,t}|q, o_{i,<t})}{\pi_{\theta_{\text{old}}}(o_{i,t}|q, o_{i,<t})}$ is the probability ratio
	, $\hat{A}_{i,t} = \frac{R_i - \text{mean}(\mathbf{R})}{\text{std}(\mathbf{R})}$ is the normalized advantage
	,$\beta$ controls the KL divergence penalty from reference policy $\pi_{\text{ref}}$.

\subsubsection{Composite Reward Function}
R$^2$GRPO's core innovation is a composite reward function for IE tasks:
\begin{equation}
	R(o_i, x, y_{\text{gold}}) = w_1 R_{\text{F1}}(o_i, y_{\text{gold}}) + w_2 R_{\text{span}}(o_i, y_{\text{gold}}) + w_3 R_{\text{relevancy}}(o_i, x) + w_4 R_{\text{rule}}(o_i, x),
\end{equation}
where $w_j$ are tunable weights and:
\begin{itemize}[leftmargin=*]
	\item \textbf{F1 Score Reward $R_{\text{F1}}$}  measures F1 score between predicted and gold extractions: $R_{\text{F1}}(o_i, y_{\text{gold}}) = \text{F1-score}(o_i, y_{\text{gold}})$.
	
	\item \textbf{Entity Span Reward} $R_{\text{span}}$ encourages precise boundaries $R_{\text{span}}(o_i, y_{\text{gold}}) = \frac{1}{N_e} \sum_{j=1}^{N_e} \text{Jaccard}(\text{span}(e_{\text{pred},j}), \text{span}(e_{\text{gold},j}))$, where $N_e$ is the number of matched entities. $\text{Jaccard}(\cdot)$~\cite{Jaccardniwattanakul2013using} measures the word-level overlap between the predicted and ground-truth entities.
	\item \textbf{Rule-pattern Reward} $R_{\text{rule}}$ rewards logical/domain pattern adherence: $R_{\text{rule}}(o_i, x) = \sum_{k} w_k \cdot \mathbb{I}(\text{pattern}_k \text{ satisfied by } o_i \text{ given } x)$. It encourage the think content to follow the rule based reasoning pattern like 'cause', 'leads to', 'rule implies', ... or explanation of the relation inference. 
	
	\item \textbf{Relevancy Reward}  $R_{\text{relevancy}}$ promotes evidence-based extraction: $R_{\text{relevancy}}(o_i, x) = \text{Map}(c_i, \text{evidence}_{\text{gold}}) - \lambda_{\text{penalty}} \cdot \left( \frac{\text{length}(c_i)}{\text{length}(x_{\text{sentence}})} \right)^2 \cdot \mathbb{I}(\text{length}(c_i) > \text{threshold})$, where $c_i$ is cited evidence. For $\text{Map}$ here, we check if the cited content appears in the original content.
	
\end{itemize}
\subsubsection{Training Strategy}
\label{subsubsec:rl_training_strategy}
To enhance efficiency, we employ:

\textbf{Curriculum Learning:} Starting with simpler IE tasks and gradually introducing complexity. We define the difficulty based on the number of entities and relation triple within one sentence.

\textbf{Data Selection:} Prioritizing instances where SFT performs poorly but clear reward signals exist. This can boost the efficiency of training of R$^2$GRPO to more epochs. The subset we selected ensure similar distribution of entity and relation types of the whole datasets and also maintain the distribution of samples with different difficulty levels.

	\subsection{Theoretical Analysis: Why Structured Reasoning Works}
	\label{subsec:mimic_theory}
	Why does hierarchical reasoning with templated steps improve performance? We provide a theoretical explanation from multiple perspectives.
	
	\textbf{Constraint Satisfaction Through Decomposition.} The hierarchical reasoning approach transforms the constrained generation problem into a more tractable form by decomposing it into stages. For MimicSFT with a single reasoning level $z_1$:
	\begin{equation}
	    P(y|x; \theta) \approx \sum_{z_1 \in \mathcal{Z}_1} P(y|z_1, x; \theta) P(z_1|x; \theta),
	\end{equation}
	where $\mathcal{Z}_1$ is the space of valid reasoning templates. This decomposition allows the model to first focus on generating valid reasoning ($z_1$) that satisfies intermediate constraints before producing the final output ($y$).
	For R$^2$GRPO with two reasoning levels:
	\begin{equation}
	    P(y|x; \theta) \approx \sum_{z_1 \in \mathcal{Z}_1} \sum_{z_2 \in \mathcal{Z}_2(z_1)} P(y|z_2, z_1, x; \theta) P(z_2|z_1, x; \theta) P(z_1|x; \theta),
	\end{equation}
	where $\mathcal{Z}_2(z_1)$ is the space of valid second-level reasoning conditioned on $z_1$. This further decomposition allows for more refined constraint satisfaction:
$z_1$ (\texttt{<reasoning>...</reasoning>}) establishes the general reasoning framework, addressing schema constraints,
$z_2$ (\texttt{<think>...</think>}) refines the reasoning with task-specific details, addressing factual constraints
,$y$ produces the final structured output based on both reasoning levels
	
	\subsubsection{Multi-Level Reasoning in R$^2$GRPO}
	R$^2$GRPO extends MimicSFT by adding a second level of reasoning optimization. If $z_1$ is the fixed reasoning template (from MimicSFT) and $z_2$ is the RL-optimized reasoning, the full generation becomes $y' = (z_1, z_2, y)$, creating a hierarchical structure:

	\textbf{Improved Constraint Satisfaction.} We can show that this hierarchical approach improves constraint satisfaction probability. Let $\mathcal{C} = \{y : C_{\text{schema}}(y) = 1 \land C_{\text{factual}}(y, x) = 1\}$ be the set of outputs satisfying all constraints. The probability of generating a valid output is:
	\begin{equation}
	    P(y \in \mathcal{C}|x; \theta) = \sum_{y \in \mathcal{C}} P(y|x; \theta).
	\end{equation}

	For the hierarchical model with reasoning steps $z_1$ and $z_2$, we assume that:
	\begin{equation}
	    P(y \in \mathcal{C}|x; \theta_{\text{hier}}) \geq P(y \in \mathcal{C}|x; \theta_{\text{direct}}),
	    \label{eq:advantage}
	\end{equation}

	when the reasoning steps are optimized to guide the model toward constraint satisfaction. 

	We will verify this later through experiments as shown in Figure~\ref{fig:bestk_max_scier} and 
	Figure~\ref{fig:bestk_ood}.
	
	\textbf{Unified Gradient Framework}
	\label{subsec:unified_gradient}
	Both SFT and RL update model parameters $\theta$ by following a gradient. Following \citet{shao2024deepseekmath}, we conceptualize these post-training algorithms under a unified gradient expression:
	\begin{equation}
		\label{eq:unified_gradient}
		\nabla_{\theta} J(\theta) = \mathbb{Eaeda}_{x, o} \left[ \frac{1}{|o|} \sum_{t=1}^{|o|} GC(x, o, t, \pi_{\text{ref}}) \nabla_{\theta} \log \pi_{\theta}(o_t | x, o_{<t}) \right],
	\end{equation}
	where $(x, o)$ is an input-output pair from distribution $\mathcal{D}$, $\pi_{\theta}(o_t | x, o_{<t})$ is the probability of generating token $o_t$ given input $x$ and previous tokens $o_{<t}$, $GC(x, o, t, \pi_{\text{ref}})$ is the gradient coefficient determining update magnitude and direction, and $\mathcal{D}$ represents the data source (human-annotated for SFT, model-generated for RL).
	
	For SFT, $GC=1$ for all tokens in the target sequence, while for RL methods like GRPO, $GC$ is derived from reward signals and advantage estimates. 
	Based on this, both SFT and GRPO can update the model parameter based on the data. Since GRPO is can refine the reasoning path and SFT(with distillation) can improve the reasoning capacity. SFT should also be able to refine the reasoning process in a simple way. And GRPO should also be able to improve the reasoning capacity and memorize knowledge from the input data.
	Our method is one step towards this.

	\section{Experiments}
	\label{sec:experiments}
	
	\subsection{Experimental Setup}
\textbf{{Training Settings}}
\textit{{Base Model}}
All our fine-tuning experiments are conducted by adapting the Qwen2.5-7B-Instruct model \cite{qwen2yang2024}.
% Qwen2.5-7B-Instruct is a powerful instruction-tuned language model known for its strong instruction following capabilities and robust baseline performance, making it a suitable candidate for investigating the efficacy of different post-training strategies for specialized IE tasks.
%	We evaluate several training settings to understand their impact on information extraction capabilities:
\textit{SFT (Supervised Fine-Tuning):} Standard fine-tuning on the target IE tasks.
\textit{MimicSFT (Multi-Task):} An SFT approach that encourages pseudo-reasoning steps and leverages multi-task learning across different IE sub-tasks (e.g., NER only, RE with Gold Entities, End-to-End IE) as described in Section \ref{subsec:sft_exploration} and \ref{subsec:sft_exploration}.
\textit{GRPO-only:} Reinforcement learning using Group Relative Policy Optimization \cite{shao2024deepseekmath} with a basic F1 score as the reward signal.
\textit{R$^2$GRPO:} Our proposed Reinforcement Learning framework, R$^2$GRPO (Relevance and Rule-Induction Group Relative Policy Optimization), incorporating a composite reward function as detailed in Section \ref{subsec:r2_rl}.
The overall prompt can be seen in the appendix~\ref{appendix:prompt}.
The system prompt for R$2$GRPO training:
\begin{tcolorbox}[
	colback=gray!10, 
	colframe=gray!80, 
	title=System Prompt,
	fonttitle=\bfseries,
	boxrule=0.3mm,
	sharp corners,
	width=\textwidth
	]
	\noindent Respond in the following format:\\
	<reasoning>\\
	Provide step-by-step reasoning to solve the task based on the given instructions and sentence.\\
	</reasoning>\\
	<think>\\
	Cite the specific sentence part (e.g., phrase, verb, or structure) supporting the relation.
	Articulate a symbolic pattern you discovered (e.g., "The verb 'achieves' suggests a Method is applied to a Task, implying a relation").
	Explain how this pattern leads to the predicted relation, referencing the relationship definition.
	Use concise, logical chains (e.g., "X performs Y $\rightarrow$ relation Z because of definition").\\
	</think>\\
	<answer>\\
	Provide the final answer in JSON format as specified in the instruction.\\
	</answer>
	\vspace{-4pt}
	\label{prompt:system}
\end{tcolorbox}

\textbf{{Implementation Details}}
	All models are fine-tuned using the LoRA approach with a rank of 16 and alpha of 32 for SFT and a rank of 64 and alpha of 128 for R$^2$GRPO, applied to all linear layers in the transformer blocks. For SFT and MimicSFT, we train for 3 epochs with a learning rate of $2 \times 10^{-5}$ and a batch size of 32 (accumulated over gradient accumulation steps). For R$^2$GRPO, the learning rate for the policy updates is set to $1 \times 10^{-6}$. More detail can be found in the appendix.
	
\textbf{{Evaluation Metrics}}
	For Named Entity Recognition (NER) and Relation Extraction (Rel and Rel+), we report the standard micro F1-score. NER: An entity is correct if its span and type match a gold entity. Rel: A relation is correct if the types and spans of both entities and the relation type match a gold relation. Rel+: It further requires the entity type is correct in the triples.

	To understand the upperbound and average performance characteristics, especially for RL-finetuned models, we employ metrics analogous to pass@K used in mathematical reasoning. We report:
	{Best F1@K:} The best F1 score among K generated outputs for a given input. This helps assess the model's capability to produce a correct extraction within its top K hypotheses. 
{Avg@K:} The average F1 score over K generated outputs, providing insight into the general quality and consistency of the model's generations.
	Unless otherwise specified, K is set to 1 for Best F1@K in main result tables. For the detailed Best F1@K analysis in Section \ref{subsec:what_reasoning_learns}, we explore a wider range of K values.

For the main results of our models, we set temperature at 0. For the baseline models we use there default setting in their documents. 
For the Best@K performance to allow better exploration, we set temperature 1.0 for all the compared models.
We show more analysis about temperature in the experiment part.

\textbf{{Baseline Models}}
We compare with:

Proprietary or large (>72B) LLMs: 
regular LLMs like Gemini2.0-flash,  DeepSeekV3
and reasoning LLMs like DeepSeek R1,Gemini2.0-flash-thinking;

Small regular LLMs(<=72B): Qwen2.5-7B-Instruct (our base model), Qwen2.5-32B-Instruct, 
Small reasoning LLMs through distillation(<=72B): deepseek-r1-distill-Qwen2.5-7B, 32B.

Supervised BERT-based models: Results from established SciBERT (Beltagy et al., 2019) fine-tuned on SciER are included for context.
General-purpose LLMs are evaluated using zero-shot.

%We also compare fine-tuned Qwen2.5-7B-Instruct models with different post-training techniques: SFT, mimicSFT, GRPO,R$^2$GRPO.

	\textbf{{Dataset}}
We conduct our experiments primarily on the SciER dataset and OOD datasets \cite{zhang2024scier}. SciER is a benchmark for information extraction in the scientific domain. It contains  24k entities and 12k relations over 106 scientific publications. It features diverse entity types (e.g., Task, Method, Datasets) and relation types (e.g., Used-For, Compare-with, Feature-Of, Evaluate-with, ...).  We use the standard splits for training. The detail dataset statistics can is shown in Table~\ref{tab:dataset_stat}.

	\subsection{Main Results}
	\label{subsec:main_results}
	We present the overall performance for Named Entity Recognition (NER) and End-to-End Relation Extraction (Rel) on the SciER test set and OOD dataset in Table \ref{tab:main_results}.
	
\begin{table*}[!t]
	\centering
	\caption{Test F1 scores of different baselines on SciER and OOD setting. ``Rel'' and ``Rel+'' represent the relation extraction under boundaries and strict evaluation, respectively. R2GRPO* is the combination of mimicSFT and R2GRPO. ''Best@5'' represent best score among the performance with 5 generations. Our training is based on Qwen2.5-7B-Instruct. }
	\vspace{-5pt}
	\setlength\tabcolsep{8.5pt}
	\begin{tabular}{lcccccc}
		\toprule
		\hline
		\multicolumn{1}{c}{} & \multicolumn{3}{c}{SciER} & \multicolumn{3}{c}{OOD} \\ \cline{2-7} 
		\multicolumn{1}{c}{\multirow{-2}{*}{Methods}} & \multicolumn{1}{l}{NER} & \multicolumn{1}{l}{Rel} & \multicolumn{1}{l}{Rel+} & \multicolumn{1}{l}{NER} & \multicolumn{1}{l}{Rel} & \multicolumn{1}{l}{Rel+} \\ \hline
		\multicolumn{7}{c}{\textit{Supervised Baselines}} \\ \hline
		PURE \cite{zhong2020frustratingly} & 81.60 & 53.27 & 52.67 & 71.99 & 50.44 & 49.46 \\
		PL-Marker \cite{ye2021packed} & 83.31 & 60.06 & 59.24 & 73.93 & 59.02 & 56.68 \\
		HGERE \cite{yan2023joint} & \underline{86.85} & \text{62.32} & \text{61.10} & \textbf{81.32} & \underline{61.31} & \underline{58.32} \\ \hline
		\multicolumn{7}{c}{\textit{Zero-Shot LLMs-based Baselines}} \\ \hline
		DeepSeek-V3 & 42.45 & 18.76 & 18.76 & 57.40 & 22.66 & 22.02 \\
		DeepSeek-R1 & 60.27 & 27.98 & 27.16 & 65.95 & 32.82 & 32.25 \\
		Gemini2.0 & 69.85 & 38.38 & 38.12 & 58.53 & 27.74 & 26.93 \\
		Gemini2.0 thinking & 61.43 & 32.30 & 31.44 & 64.75 & 30.62 & 30.33 \\
		Qwen2.5-32B & 56.67 & 17.10 & 17.10 & 36.85 & 8.72 & 8.72 \\
		DeepSeek-R1-Distill-Qwen-32B & 57.63 & 17.62 & 17.11 & 49.00 & 10.79 & 9.98 \\
		Qwen2.5-7B & 41.24 & 7.09 & 7.09 & 44.88 & 4.20 & 4.20 \\
		DeepSeek-R1-Distill-Qwen-7B & 32.01 & 4.60 & 4.60 & 30.25 & 2.88 & 2.88 \\ \hline
		\multicolumn{7}{c}{\textit{Fine-tuned LLMs}} \\ \hline
		SFT & 80.76 & 42.22 & 41.01 & 70.13 & 19.45 & 18.12 \\
		GRPO & 76.18 & 48.84 & 48.02 & 68.93 & 42.34 & 41.76 \\\hline
		\multicolumn{7}{c}{\textit{Ours}} \\ \hline
		Mimic-SFT & 81.70 & 56.02 & 55.34 & 73.71 & 50.74 & 49.95 \\
		R2GRPO & 77.55 & 54.59 & 53.65 & 70.05 & 45.72 & 44.67 \\
		R2GRPO* & {84.36} & \underline{66.81} & \underline{65.95} & {77.84} & 55.08 & 54.29 \\ 
		R2GRPO*(Best@5) & \textbf{{88.90}} & \textbf{74.38} & \textbf{74.03} & \underline{{81.27}} & \textbf{66.74 }& \textbf{64.12} \\
		\hline
		\bottomrule
	\end{tabular}%
	\label{tab:main_results}
	\vspace{-10pt}
\end{table*}
	
	The results in Table \ref{tab:main_results} show that our R$^2$GRPO boosts the performance of Qwen2.5-7B-Instruct on SciER and OOD for both NER and RE tasks significantly. Especially on Relation extraction in SciER, it outperforms all the supervised baselines. Mimic-SFT achieve higher relation extraction score than SFT one show the pseudo CoT can activate model's 'reasoning' ability or constrained generation refine the reasoning path. 
	Not we use 0 temperature for our models and default setting for the baseline models.
	For the results for supervised baselines, we use the reported results from the original paper \cite{zhang2024scier}.
	Similarly, R$^2$GRPO outperform GRPO is this case.
	 MimicSFT also shows strong performance, often outperforming standard SFT, highlighting the benefit of the proposed structured pseudo-reasoning.
\subsection{What Do Reasoning Models Learn? Analysis of Best F1@K}
\label{subsec:what_reasoning_learns}
To delve deeper into what reasoning models learn, particularly through Reinforcement Learning with Verifiable Rewards (RLVR) like R$^2$GRPO, we analyze the Best F1@K performance. We selected a subset of 50 challenging samples from the SciER test set and evaluated model outputs with K values ranging from 1, 4, 16, 32, 64, 128, 512, up to 1024. This analysis aims to understand the upper-bound capabilities of the models and how SFT and RLVR shape their knowledge and reasoning. The results for NER and RE are visualized in Figure \ref{fig:bestk_max_scier}.

\textbf{RLVR and SFT both Enhance Reasoning Capacity:} From Figure~\ref{fig:bestk_max_scier} and Figure~\ref{fig:bestk_ood}, both RLVR-based (GRPO, R$^2$GRPO) and SFT-based models (SFT, MimicSFT) significantly outperform the base Qwen2.5-7B-Instruct model across all K values. This contradicts the hypothesis that RLVR merely optimizes path selection without improving underlying capabilities \cite{yue2025does}. Instead, our results demonstrate that both SFT and RLVR enable models to acquire domain-specific knowledge and enhance reasoning capabilities relevant to IE tasks. The consistent improvement in Best F1@K scores, even at large K values, indicates a genuine expansion of the model's knowledge boundaries rather than just better prioritization of existing knowledge.

\textbf{Hierarchical Reasoning Improves Knowledge Integration:} MimicSFT consistently outperforms standard SFT, and similarly, R$^2$GRPO outperforms basic GRPO across both in-domain and OOD settings. This validates our theoretical analysis in Section \ref{subsec:mimic_theory} that structured decomposition of reasoning facilitates better constraint satisfaction. The templated reasoning approach creates intermediate representations that guide the model toward valid outputs, effectively narrowing the search space while maintaining exploration capabilities.
This verify our assumption in Eq.~\ref{eq:advantage}.
 
 \begin{wrapfigure}{r}{0.48\textwidth}
 	\centering
 	\includegraphics[width=0.95\linewidth]{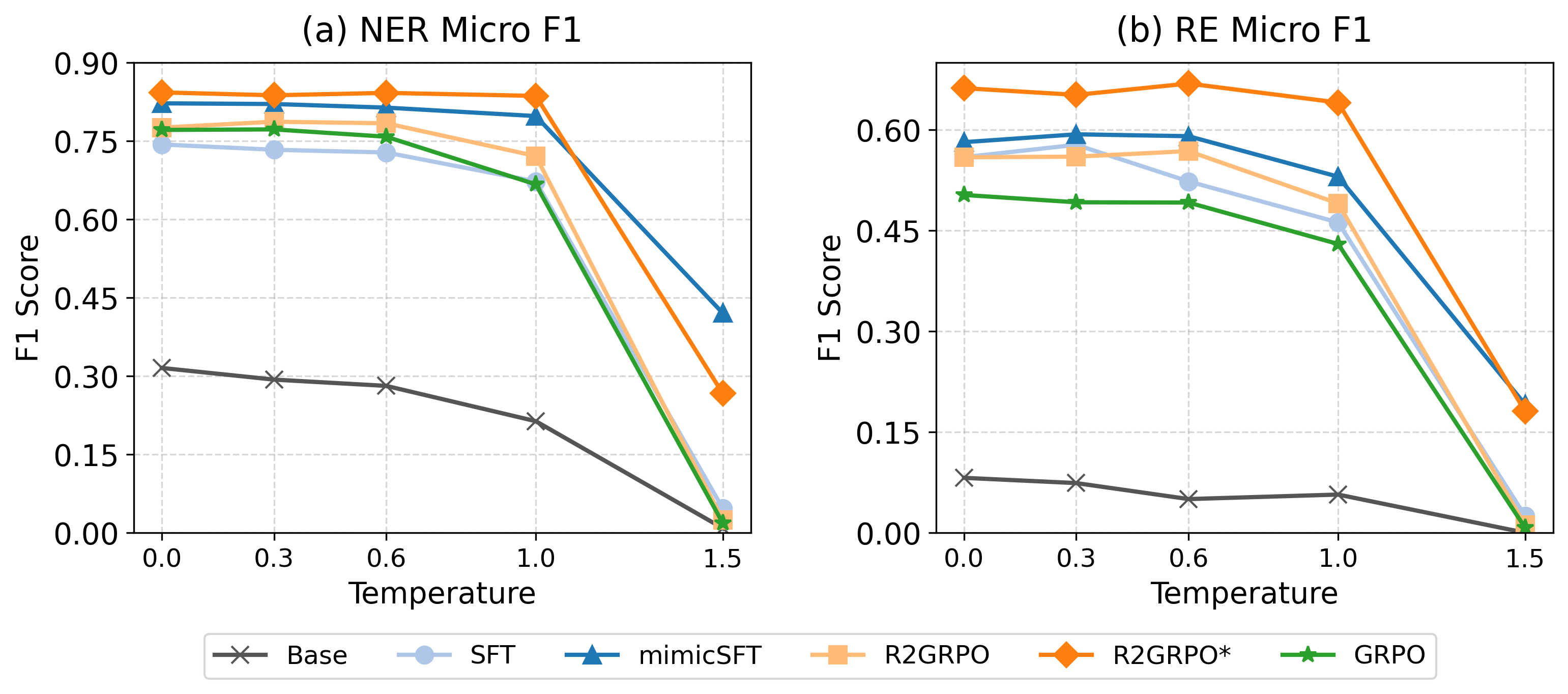}
 	\vspace{-5pt}
 	\caption{Performance v.s temperature}
 	\label{fig:performance_temperature}
 	\vspace{-1pt}
 \end{wrapfigure}

\textbf{Complementary Effects of SFT and RLVR:} While SFT-based models (particularly MimicSFT) achieve slightly higher Best F1@K at very large K values, RLVR models demonstrate superior Avg@K and Best F1@1 scores. This reveals a fundamental trade-off: SFT expands the model's knowledge boundaries, while RLVR optimizes the probability distribution to prioritize high-quality outputs. The combination in R$^2$GRPO* achieves the best of both worlds—maintaining high Best F1@K (knowledge breadth) while significantly improving Best F1@1 (practical performance).

\textbf{Structured Reasoning Enhances Generalization:} The performance gap between our methods and baselines widens in OOD settings (Figure~\ref{fig:bestk_ood}), demonstrating that hierarchical reasoning improves generalization. This aligns with our theoretical framework—by decomposing complex extraction tasks into structured sub-problems, models learn more generalizable patterns rather than memorizing specific input-output mappings. The structured attention allocation mechanism described in Section \ref{subsec:mimic_theory} enables more effective feature extraction across different domains.
\begin{wrapfigure}{r}{0.49\textwidth}
	\centering
	\subfloat[Response length]{\includegraphics[width=0.23\textwidth]{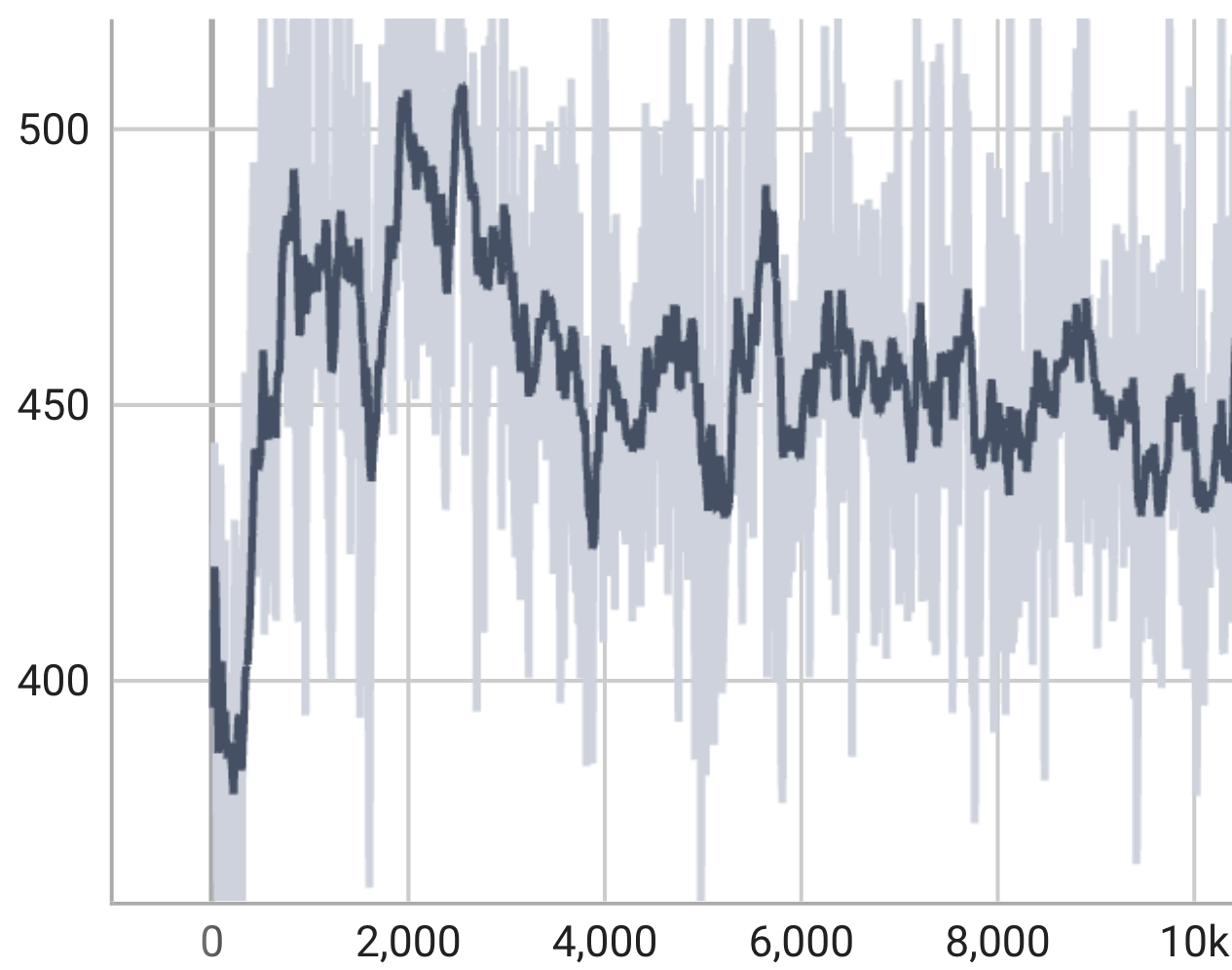}}\quad
	\subfloat[Reward]{\includegraphics[width=0.23\textwidth]{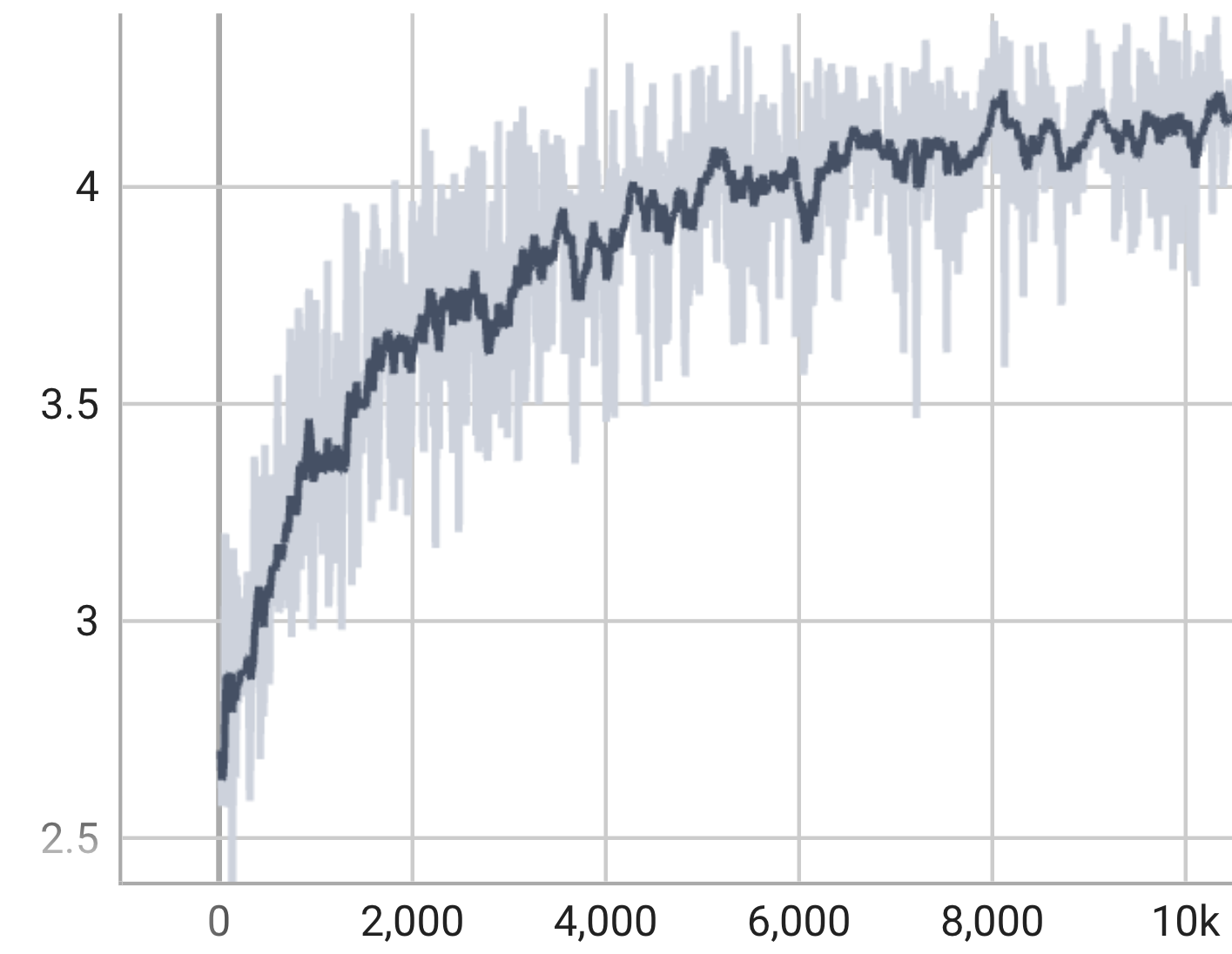}}
	\vspace{-0pt}
	\caption{Response length(a) and Reward(b) v.s. training steps for R$^2$GRPO}
	\label{fig:grpo_train}
\end{wrapfigure}

\textbf{Exploration-Exploitation Balance:} The slightly lower Best F1@K of R$^2$GRPO compared to MimicSFT at very large K values reflects an intentional trade-off. R$^2$GRPO optimizes for high-reward trajectories within a practical exploration budget, focusing computational resources on promising reasoning paths. This is particularly valuable in real-world applications where generating hundreds of candidates is impractical. The higher Avg@K scores of R$^2$GRPO indicate more consistent performance across generations, making it more reliable in production environments.

\vspace{-4pt}
\subsubsection{Ablation Studies and Parameter Sensitivity}
\begin{wrapfigure}{r}{0.49\textwidth}
	\centering
	\includegraphics[width=0.96\linewidth]{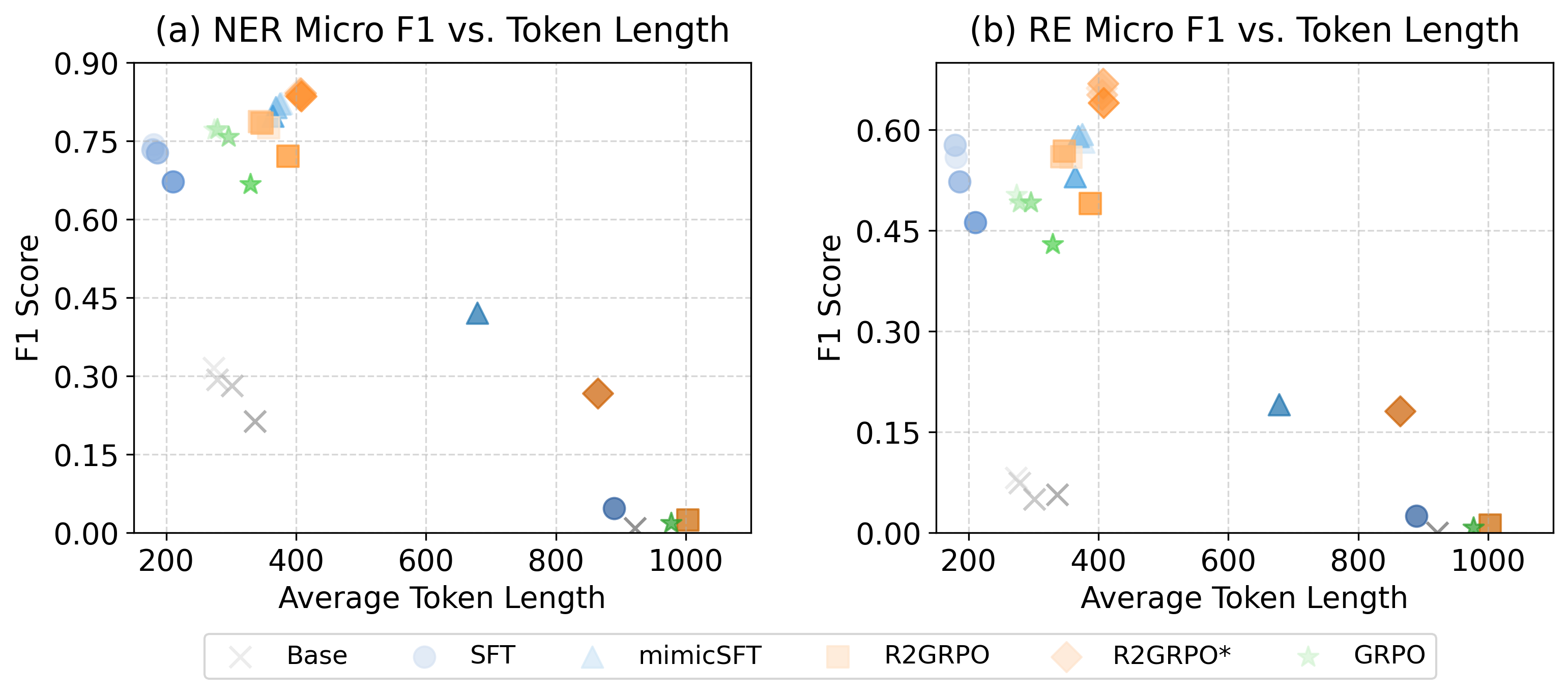}
	\vspace{-2pt}
	\caption{Performance v.s response token length. The deeper the color, the higher the temperature.}
	\label{fig:performance_length}
\end{wrapfigure}

\textbf{Component Contribution Analysis:} Table~\ref{tab:main_results} demonstrates the progressive improvement from SFT to MimicSFT and from GRPO to R$^2$GRPO, validating each component's contribution. The most substantial gains come from combining MimicSFT with R$^2$GRPO (R$^2$GRPO*), which achieves a 24.59 point improvement in Rel F1 over standard SFT and a 17.97 point improvement over basic GRPO. This synergistic effect confirms our hypothesis that knowledge acquisition (primarily through SFT) and reasoning refinement (primarily through RL) are complementary processes that can be jointly optimized.

\textbf{Temperature Sensitivity:} Figure~\ref{fig:performance_temperature} shows that our models consistently outperform baselines across different temperature settings.  The optimal performance is at lower temperatures (<0.6). This indicates that SciIE benefits from more deterministic generation since the task requires precise entity boundary detection and relation inference based on the content. 
The performance degradation at higher temperatures suggests that excessive exploration introduces more noise in the structured extraction process.
We also found that the completion length increases with higher temperature in Figure~\ref{fig:Completion Length}.
This suggests the thinking length increase with the noise that leads to unstable thinking content can harm the performance.

\textbf{Response Length Analysis:} In Figure~\ref{fig:grpo_train}, as training goes, the response length first increase than decrease. This differs from the results of \cite{shao2024deepseekmath} where response length increase.
 Figure~\ref{fig:performance_length} reveals an interesting relationship between response length and performance. The performance does not generally benefit from longer response. The long response length at high temperature(1.5) leads to bad performance. This suggest for tasks like SciIE, the constrained reasoning process is better than the long but noisy think content.
 This supports our assumption—effective IE requires concise, targeted reasoning that focuses on relevant constraints rather than exhaustive exploration. The hierarchical reasoning approach in R$^2$GRPO guides the model to generate more efficient reasoning paths, avoiding unnecessary elaboration while maintaining extraction accuracy. 
More training detail is in Figure~\ref{fig:training_detail}.
\vspace{-5pt}
\section{Limitation and Future Work}
Our study mainly focuses on the SciIE. The effectiveness of MimicSFT's pseudo-reasoning templates and R$^2$GRPO might vary across different types of information extraction tasks or languages. In the future, we would explore the adaptability of R$^2$GRPO to broader domains and investigate more automated methods for generating or refining reasoning templates.  Moreover, further research can also focus on how structured reasoning influences knowledge acquisition and path selection in diverse LLM architectures and explore the scalability of our approach to even larger models or datasets of different domains.

\bibliographystyle{plainnat}  
	\bibliography{references} 
	\newpage
\appendix
\section{Appendix}
\label{appendix}
\subsection{R$^2$GRPO Training}

\subsection{Dataset Statistics}
We show here the detail statistic of the datasets used. For the Best@K evaluation, we select 50 samples from both the SciER and OOD test set.
And for the rest evaluation we use the full datasets.
\begin{table}[h!]
	\centering
	\caption{Detail Distribution of Datasets}
	{
		\begin{tabular}{lccccc}
			\toprule
			\textbf{Entity/Relation Type} & \textbf{Train} & \textbf{Dev} & \textbf{SciER Test} & \textbf{OOD Test} & \textbf{Total} \\ 
			\midrule
			\multicolumn{6}{c}{\textbf{Entity Types}} \\ 
			\midrule
			\texttt{Method} & 11424 & 1549 & 1890 & 1018 & 15881 \\
			\texttt{DATASET} & 3220 & 269 & 370 & 83 & 3942 \\
			\texttt{TASK} & 3397 & 416 & 688 & 194 & 4695 \\
			\textbf{Total} & 18041 & 2234 & 2948 & 1295 & 24518 \\ 
			\midrule
			\multicolumn{6}{c}{\textbf{Relation Types}} \\ 
			\midrule
			\texttt{PART-OF} & 1865 & 214 & 304 & 111 & 2494 \\
			\texttt{USED-FOR} & 2398 & 343 & 546 & 167 & 3454 \\
			\texttt{EVALUATED-WITH} & 863 & 78 & 131 & 49 & 1121 \\
			\texttt{SYNONYM-OF} & 880 & 76 & 170 & 89 & 1215 \\
			\texttt{COMPARE-WITH} & 875 & 175 & 114 & 54 & 1218 \\
			\texttt{SUBCLASS-OF} & 697 & 114 & 176 & 73 & 1060 \\
			\texttt{BENCHMARK-FOR} & 551 & 64 & 85 & 28 & 728 \\
			\texttt{SUBTASK-OF} & 210 & 31 & 65 & 9 & 315 \\
			\texttt{TRAINED-WITH} & 404 & 37 & 35 & 2 & 478 \\ 
			\textbf{Total} & 8743 & 1132 & 1626 & 582 & 12083 \\
			\bottomrule
		\end{tabular}
	}
	\label{tab:dataset_stat}
\end{table}

\subsection{Training details}
\begin{figure}[!t]
	\centering
	\subfloat[Completion Length]{
		\label{fig:Completion Length}
		\includegraphics[width=0.3\textwidth]{figures/length}}
	\subfloat[Ner reward]{
		\label{fig:nerreward}
		\includegraphics[width=0.3\textwidth]{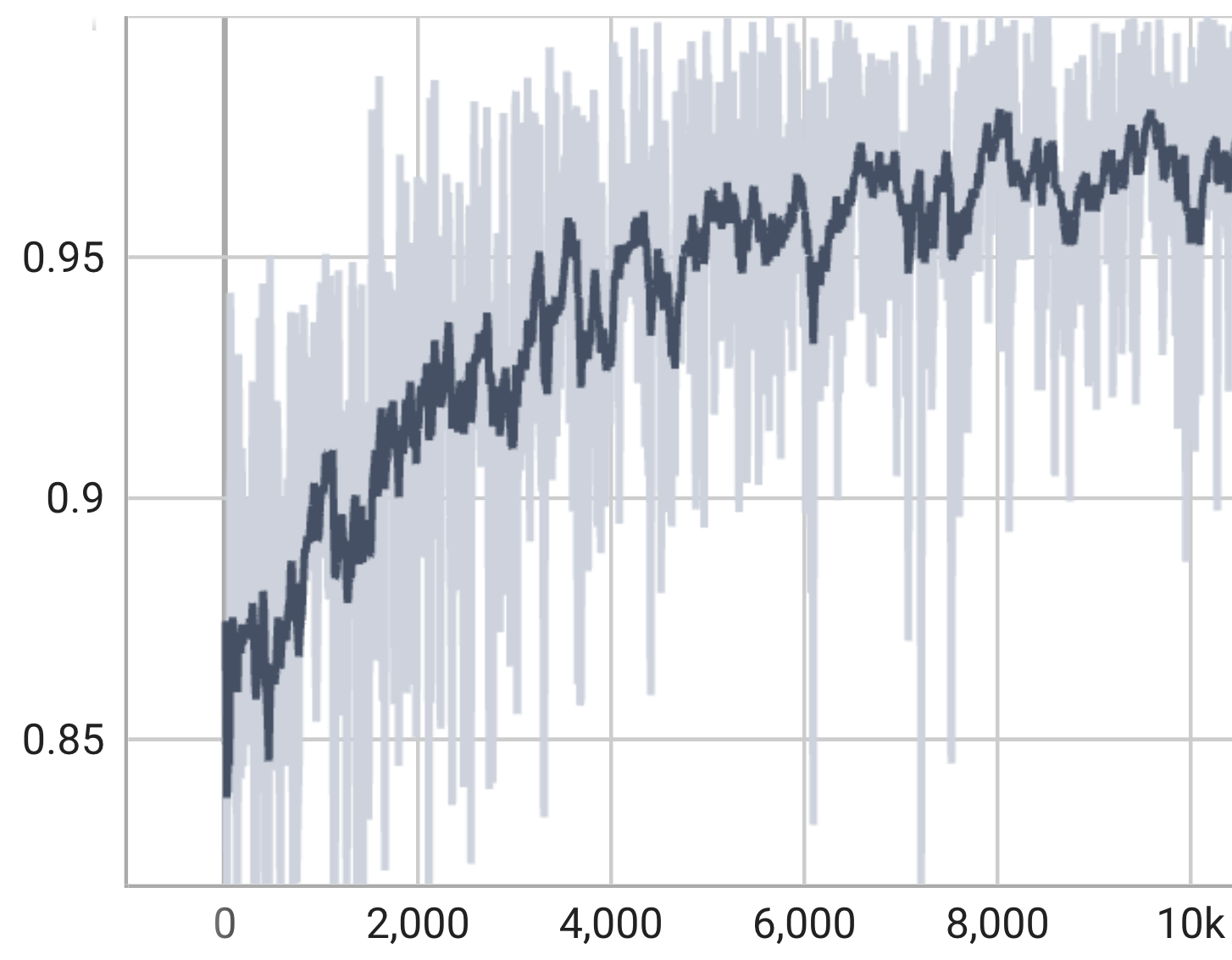}}
	\subfloat[Rel reward]{
		\label{fig:relreward}
		\includegraphics[width=0.3\textwidth]{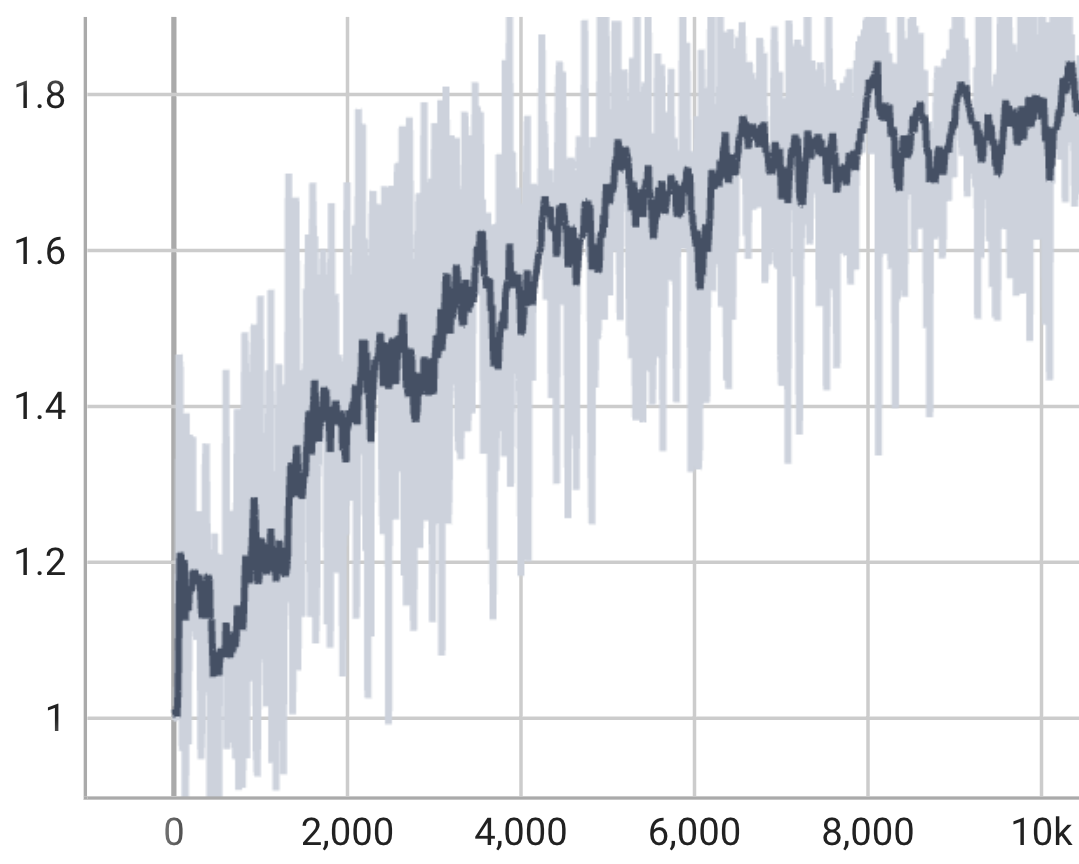}}
	\\
	\subfloat[Reasoning reward]{
		\label{fig:reasonreward}
		\includegraphics[width=0.3\textwidth]{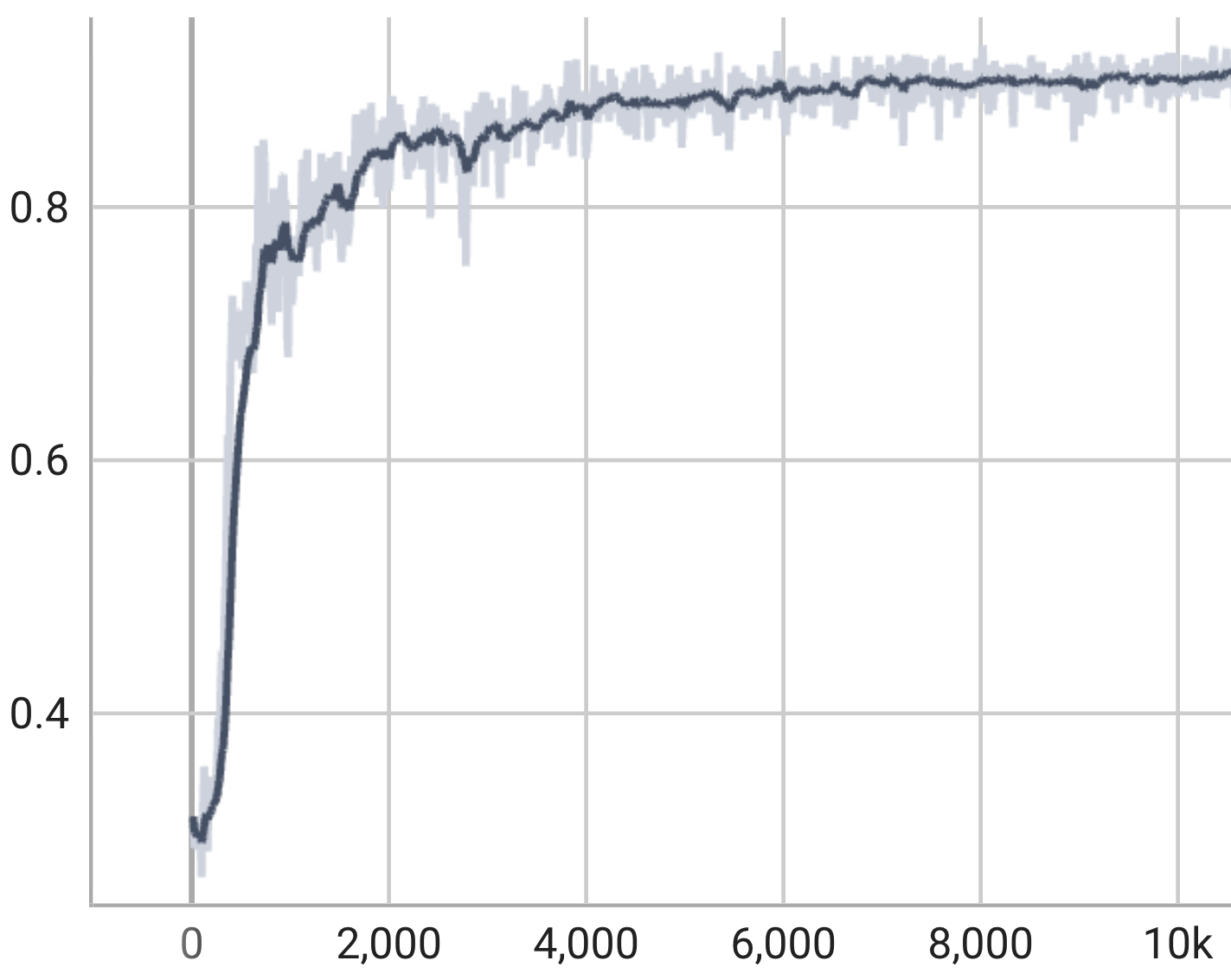}}
	\subfloat[Total reward]{
		\label{fig:reward}
		\includegraphics[width=0.3\textwidth]{figures/reward}}
	\subfloat[Standard Deviation]{
		\label{fig:std}
		\includegraphics[width=0.3\textwidth]{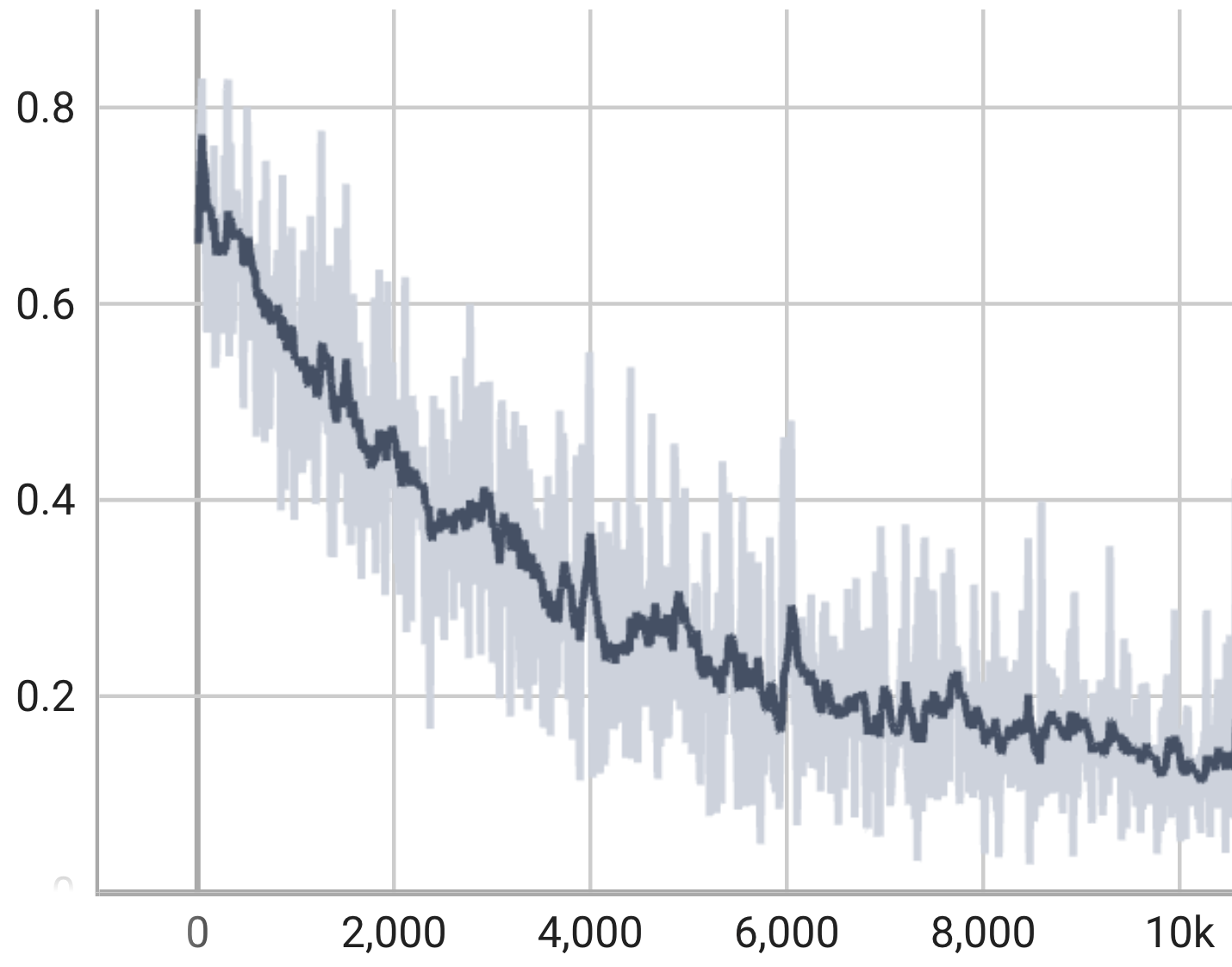}}
	\caption{
		R$^2$GRPO training detail v.s. steps
	}
	\label{fig:training_detail}
\end{figure}

We trained the model on a sub set of the SciER with 1K sample for RL. The selection is based on the balanced distribution of the entity and relation of different types and samples with different length total number of entities and relation triples. The training detail is shown in Fig.~\ref{fig:training_detail}.

Training can be done within 24GB vram with the lora adapter. However, larger group size require larger vram
We train for 3 epochs on the full SciER datasets and 10 epochs for the RL on the 1K subset.

\subsection{Prompt}
\label{appendix:prompt}
We adapt the instruction from SciER \cite{zhang2024scier}.
\begin{tcolorbox}[
	colback=gray!10, % Light gray background
	colframe=gray!80, % Dark gray border
	title=Ner Background,
	fonttitle=\bfseries,
	boxrule=0.5mm,
	sharp corners,
	width=\textwidth
	]
	
	Extract specific entities from the following sentence. The entities to be identified are: 'Dataset', 'Task', and 'Method'.

\#\#\# Entity Definitions:

- 'Task': A task in machine learning refers to the specific problem or type of problem that a ML/AI model/method is designed to solve. Tasks can be broad, like classification, regression, or clustering, or they can be very specific, such as Pedestrian Detection, Autonomous Driving, Sentiment Analysis, Named Entity Recognition, and Relation Extraction.

- 'Method': A method entity refers to the approach, algorithm, or technique used to solve a specific task/problem. Methods encompass the computational algorithms, model architectures, and the training procedures that are employed to make predictions or decisions based on data. For example, Convolutional Neural Networks, Dropout, data augmentation, recurrent neural networks.

- 'Dataset': A realistic collection of data that is used for training, validating, or testing the algorithms. These datasets can consist of various forms of data such as text, images, videos, or structured data. For example, MNIST, COCO, AGNews, IMDb.

\#\#\# Other Notes:
- Generics cannot be used independently to refer to any specific entities, e.g., 'This task', 'the dataset', and 'a public corpus' are not entities.

- The determiners should not be part of an entity span. For example, given span 'the SQuAD v1.1 dataset', where the determiner 'the' should be excluded from the entity span.

- If both the full name and the abbreviation are present in the sentence, annotate the abbreviation and its corresponding full name separately. For instance, '20-newsgroup (20NG)'.

- Only annotate "factual, content-bearing" entities. Task, dataset, and method entities normally have specific names and their meanings are consistent across different papers. For example, "CoNLL03", "SNLI" are factual entities. Annotators should annotate only the minimum necessary to represent the original meaning of task/dataset/metric (e.g., "The", "dataset", "public", 'method', 'technique' are often omitted).

Based on the given sentence and the entities with their types, determine the relationship between each pair. The potential relations are: ['Part-Of', 'SubClass-Of', 'SubTask-Of', 'Benchmark-For', 'Trained-With', 'Evaluated-With', 'Synonym-Of', 'Used-For', 'Compare-With']. If no relationship exists between a pair, do not include it in the output.
	
	\vspace{-4pt}
	\label{prompt: ner}
\end{tcolorbox}

\begin{tcolorbox}[
	colback=gray!10, % Light gray background
	colframe=gray!80, % Dark gray border
	title=Rel Background,
	fonttitle=\bfseries,
	boxrule=0.5mm,
	sharp corners,
	width=\textwidth
	]
	\#\#\# Relationship Definitions:
	
	- 'Part-Of': This relationship denotes that one entity (e.g., a Method) is a component or a part of another entity (e.g., another Method).
	
	- 'SubClass-Of': Specifies that one entity is a subclass or a specialized version of another entity.
	
	- 'SubTask-Of': Indicates that one Task is a subset or a specific aspect of another broader Task.
	
	- 'Benchmark-For': Shows that a Dataset serves as a standard or benchmark for evaluating the performance of a Method on a Task.
	
	- 'Trained-With': Indicates that a Method is trained using a Dataset.
	
	- 'Evaluated-With': This relationship denotes that a Method is evaluated using a Dataset to test its performance or conduct experiments.
	
	- 'Synonym-Of': Indicates that two entities are considered to have the same or very similar meaning, such as abbreviations.
	
	- 'Used-For': Shows that one entity (e.g., a Method) is utilized for achieving or performing another entity (e.g., a Task). This relationship is highly flexible.
	
	- 'Compare-With': This relationship is used when one entity is compared with another to highlight differences, similarities, or both.
	
	\#\#\# Notes:
	
	- Determine the 'Relationship' that best describes how the subject and object are related, based on the sentence context.
	
	- Please do not annotate negative relations (e.g., X is not used in Y).
	
	- Annotate a relationship only if there is direct evidence or clear implication in the text. Avoid inferring relationships that are not explicitly mentioned or clearly implied.
	\vspace{-4pt}
	\label{prompt: rel}
\end{tcolorbox}

\begin{tcolorbox}[
	colback=gray!10, % Light gray background
	colframe=gray!80, % Dark gray border
	title=Task,
	fonttitle=\bfseries,
	boxrule=0.5mm,
	sharp corners,
	width=\textwidth
	]
	Given the sentence: "{sentence}"
	
	Extract entities and their relations.
	
	\#\#\# Instruction:
	
	- Think step-by-step to identify entities ('Dataset', 'Task', 'Method') and their relationships.
	
	- Return the results in JSON format with:
	
	- "ner": a list of [entity, type] pairs.
	
	- "rel": a list of [subject, relation, object] triples.

	\vspace{-4pt}
	\label{prompt: format}
\end{tcolorbox}

In general, the prompt consists of the background definition of the entity, relation and the instruction on the tasks.

\end{document}